\documentclass[11pt]{article}

\usepackage[preprint]{acl}

\usepackage{times}
\usepackage{latexsym}

\usepackage[T1]{fontenc}

\usepackage[utf8]{inputenc}

\usepackage{microtype}

\usepackage{inconsolata}

\usepackage{graphicx}
\usepackage{pifont}    
\usepackage{amssymb}   
\usepackage{makecell}  
\usepackage{booktabs}  
\usepackage{multirow}
\usepackage{amsmath}
\usepackage{algpseudocode}
\usepackage{amssymb}
\usepackage{xcolor}
\usepackage{siunitx}
\usepackage{algorithm}
\usepackage{newfloat}
\usepackage{listings}
\usepackage{array}
\usepackage{hyperref}
\usepackage{fontawesome5}
\title{ADAPT: Benchmarking Commonsense Planning under Unspecified Affordance Constraints}

\author{
  \textbf{Pei-An Chen\textsuperscript{1}},
  \textbf{Yong-Ching Liang\textsuperscript{1}},
  \textbf{Jia-Fong Yeh\textsuperscript{1}},
  \textbf{Hung-Ting Su\textsuperscript{1}},
  \textbf{Yi-Ting Chen\textsuperscript{2}},
  \textbf{Min Sun\textsuperscript{3}},
  \textbf{Winston H. Hsu\textsuperscript{1}}
\\
  \textsuperscript{1}National Taiwan University \\
  \textsuperscript{2}National Yang Ming Chiao Tung University \\
  \textsuperscript{3}National Tsing Hua University
\\
  \small{
    \faEnvelope\ : \href{mailto:charlottechen@cmlab.csie.ntu.edu.tw}{charlottechen@cmlab.csie.ntu.edu.tw}
  \quad
  \faGlobe\ : \url{https://charlotteannchen.github.io/ADAPT/}
  }
}

\begin{document}
\maketitle

\begin{abstract}
Intelligent embodied agents should not simply follow instructions, as real-world environments often involve unexpected conditions and exceptions. However, existing methods usually focus on directly executing instructions, without considering whether the target objects can actually be manipulated, meaning they fail to assess available affordances. To address this limitation, we introduce \textbf{DynAfford}, a benchmark that evaluates embodied agents in dynamic environments where object affordances may change over time and are not specified in the instruction. DynAfford requires agents to perceive object states, infer implicit preconditions, and adapt their actions accordingly. To enable this capability, we introduce \textbf{ADAPT} (Affordance-Driven Adaptive Planning and Task execution), a plug-and-play module that augments existing planners with explicit affordance reasoning. Experiments demonstrate that incorporating ADAPT significantly improves robustness and task success across both seen and unseen environments. We also show that a domain-adapted, LoRA-finetuned vision-language model used as the affordance inference backend outperforms a commercial LLM (GPT-4o), highlighting the importance of task-aligned affordance grounding.
\end{abstract}

\begin{figure*}[tb]
  \centering
  \includegraphics[width=1.00\linewidth]{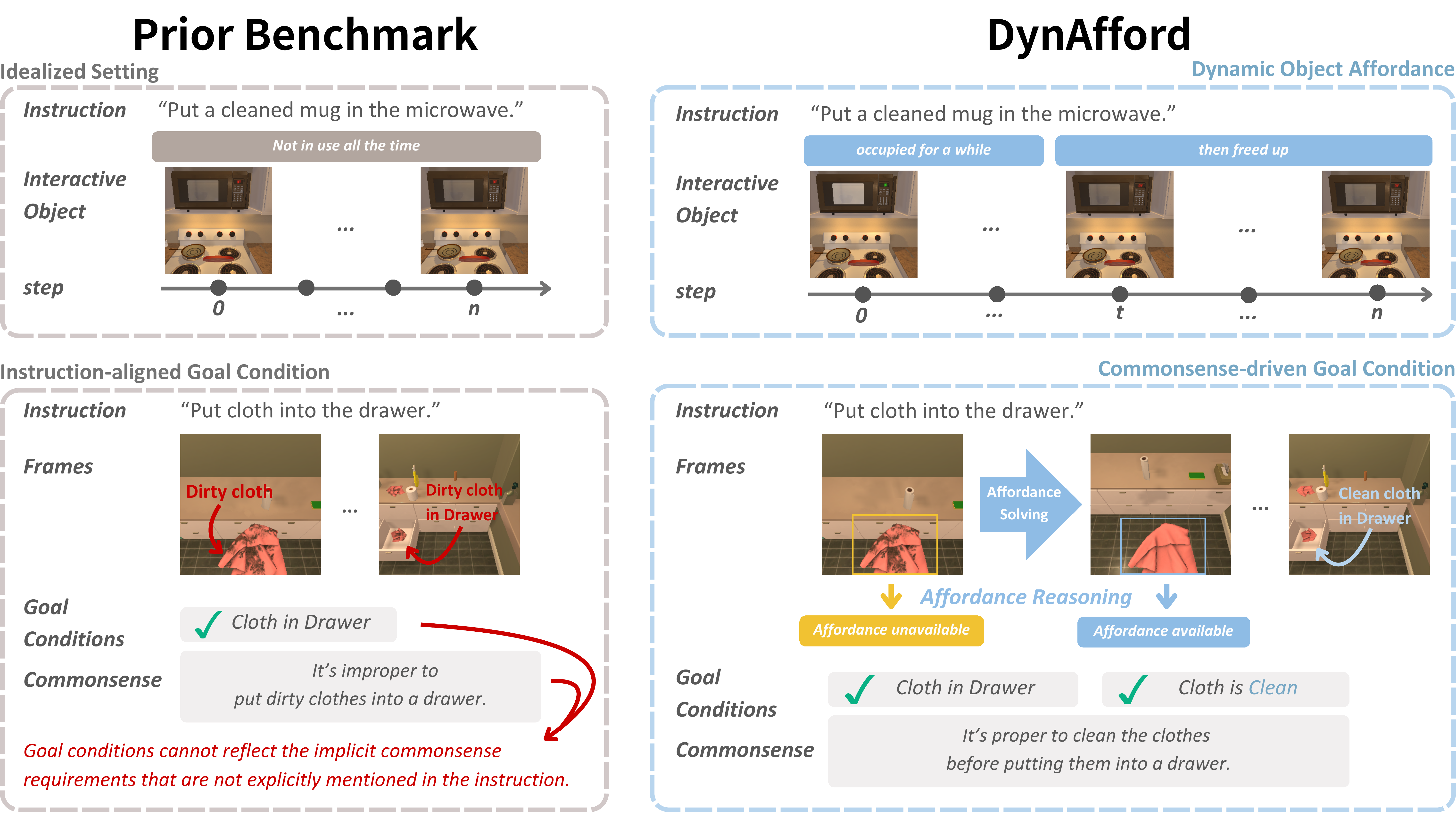}
  \caption{
  \textbf{Overview of DynAfford benchmark.} Unlike prior embodied benchmarks (left) that assume static object usability and fully specified goals, DynAfford introduces dynamic object affordances and commonsense-driven goal conditions (right). Agents must detect latent preconditions (e.g., cleanliness), resolve temporarily inapplicable actions, and adapt their behavior beyond literal instruction following.}
  \label{fig:dataset}
\end{figure*}  

\section{Introduction}
Humans have the ability to handle unexpected scenarios that are not specified in the instructions when performing everyday tasks. For example, when instructed to put cloth into a drawer, a person encountering a dirty cloth would naturally recognize that placing it into a drawer is inappropriate. Therefore, they go beyond literal instruction-following by considering context-specific preconditions, such as cleaning the cloth before use. This ability to reason about context-dependent object usability and to infer unstated preconditions is fundamental to human commonsense reasoning, and is essential for enabling embodied agents to operate robustly in real-world household environments. It motivates this work, aiming to verify \textit{whether embodied methods can handle situations where object usability depends on state changes or surrounding conditions that are not explicitly mentioned in the instruction}.

Unfortunately, this limitation cannot be adequately studied using existing embodied AI benchmarks, 
which typically assume static object usability and fully specified goal conditions. 
As a result, agents are not required to detect unmet preconditions or reason about evolving object usability, 
which are essential for realistic deployment. In contrast, we focus on long-horizon tasks requiring dynamic affordance reasoning.

We introduce \textbf{DynAfford}, a new embodied AI benchmark designed to evaluate agents in dynamic and under-specified environments, where object affordances—treated as latent preconditions governing action applicability—may change over time and are not explicitly specified in the instruction (Figure~\ref{fig:dataset}). DynAfford requires agents to infer and resolve such implicit constraints during execution.

We evaluate several state-of-the-art methods, including MOCA \cite{singh2021factorizing}, FILM \cite{min2021film}, CAPEAM \cite{kim2023context}, LLM-Planner \cite{song2023llm}, and SayCan \cite{ahn2022saycan}, and observe substantial performance degradation under DynAfford’s dynamic settings, even when equipped with strong vision-language backends such as GPT-4o. This reveals a fundamental mismatch between existing approaches and the demands of dynamic embodied environments.

To address this gap, we propose \textbf{Affordance-Driven Adaptive Planning and Task execution(ADAPT)}, a unified decision-time inference module that jointly infers object affordance states and determines executable actions under dynamic constraints. When integrated with strong embodied agents such as FILM and CAPEAM, ADAPT significantly improves robustness, yielding up to 73.2\% relative improvements in success rate and 34.7\% in goal completion on the test unseen split.

Our contributions are summarized as follows:
\begin{itemize}
    \item We identify a key failure mode in embodied planning, namely \textbf{instruction underspecification}, where high-level instructions do not explicitly capture the conditions required for successful execution.
    \item We introduce \textbf{DynAfford}, a benchmark designed to systematically expose these limitations by incorporating dynamic affordances and commonsense-driven goal conditions, providing a realistic testbed for evaluating embodied agents under execution-time uncertainty.
    \item To address this challenge, we propose \textbf{ADAPT}, a plug-and-play module for affordance-aware planning that infers action applicability under latent preconditions, leading to substantial improvements in robustness and task success across multiple state-of-the-art agents.
\end{itemize}

\begin{table*}[t]
  \centering
  \small
  \setlength{\tabcolsep}{3pt}
  \begin{tabular}{l c c c c}
    \toprule
    \textbf{Benchmark} & 
    \textbf{Simulator} &
    \textbf{Reasoning Capability} &
    \textbf{Goal Representation} &
    \textbf{Task Specification} \\
    \midrule
    SAPien \cite{xiang2020sapien}       
      & SAPien      
      & --      
      & --      
      & User-defined \\

    VirtualHome \cite{puig2018virtualhome} 
      & Unity       
      & LH      
      & --      
      & Natural language \\

    BEHAVIOR-1K \cite{li2023behavior}  
      & OmniGibson 
      & LH + IP      
      & Symbolic (PDDL)      
      & PDDL \\

    ALFRED \cite{shridhar2020alfred}    
      & AI2-THOR    
      & LH + IP      
      & Instruction-aligned      
      & Natural language \\
    \midrule
    \textbf{DynAfford (Ours)} 
      & AI2-THOR    
      & \textbf{LH + IP + DS + CU}      
      & \textbf{Commonsense-driven}      
      & \textbf{Natural language} \\
    \bottomrule
  \end{tabular}
  \caption{
    \textbf{Comparison of major embodied AI benchmarks.}
    DynAfford uniquely supports dynamic affordance reasoning, including implicit preconditions and context-sensitive object usability under long-horizon tasks. LH: Long-horizon tasks; IP: Implicit preconditions; DS: Dynamic state transitions; CU: Context-sensitive usability.
  }
  \label{tab:dataset_comparison}
\end{table*}

\section{Related Work}

\subsection{Embodied Instruction Following}
Embodied Instruction Following (EIF) requires agents to interpret natural language or symbolic specifications such as PDDL, and execute long-horizon plans in household environments through navigation and manipulation. Benchmarks such as ALFRED \cite{shridhar2020alfred} and BEHAVIOR-1K \cite{li2023behavior} advance this area by combining perception, language, and imitation learning. Recent methods improve generalization through modular or hierarchical architectures: MOCA \cite{singh2021factorizing} decouples object grounding from action prediction; FILM \cite{min2021film} and HLSM \cite{hlsm} decompose instructions into perception, planning, and memory; ET \cite{Pashevich_2021_ICCV} and CAPEAM \cite{kim2023context} enhance temporal consistency via recurrent memory modules; HiTUT \cite{zhang2021hierarchical} models hierarchical task structures by combining subgoal planning, navigation, and manipulation using unified transformers.

Despite strong performance on existing benchmarks, most embodied instruction-following methods assume static object usability and fully specified goal conditions. A related concurrent work, VLN-NF~\cite{su2026vln}, studies false-premise instructions in which the target is absent. By contrast, we assume the target exists and instead study \textit{action infeasibility under dynamic affordance constraints}. The two works therefore address different failure modes: goal invalidity versus action infeasibility.

\subsection{LLM- and VLM-Based Grounding in Embodied Robotics}
Recent work has increasingly leveraged large language models (LLMs) and vision-language models (VLMs) to enable flexible, commonsense-driven planning for embodied agents. These approaches typically integrate language priors with visual grounding to guide action selection, either through affordance scoring or in-context planning mechanisms \cite{ahn2022saycan, brown2020language, song2023llm, chuang2018learning}. 
Such designs allow agents to generalize across tasks by leveraging large-scale pretrained knowledge and structured reasoning over action spaces.

Beyond passive perception, recent work explores active knowledge acquisition under partial observability, such as ActiveVOO \cite{liuactivevoo}, which improves object identification through active sensing.

However, these approaches primarily focus on perception, while DynAfford addresses action execution under dynamic affordance constraints, where failures arise from unmet or evolving preconditions. 
These directions are complementary.

\subsection{Affordance Reasoning in Robotics}
Robust embodied agents must reason about whether actions are feasible given the latent and dynamic properties of objects. 
Most existing approaches model affordances based on static, visually grounded attributes such as object presence or geometry \cite{ahn2022saycan}, or incorporate reasoning through language or code representations \cite{huang2022inner, logeswaran-etal-2024-code}. 
Other work explores semantic and socially grounded affordances, predicting and explaining action appropriateness from static visual scenes \cite{chuang2018learning}.

However, these approaches largely treat affordances as static or immediately observable, without accounting for temporally evolving or implicit preconditions. 
In contrast, DynAfford evaluates whether agents can reason about when actions should not be executed due to unmet or evolving preconditions, serving as a diagnostic benchmark for affordance-aware action selection under dynamic constraints.

\section{The DynAfford Benchmark}
We introduce DynAfford, a benchmark designed to evaluate embodied agents’ ability to reason compositionally and handle dynamic, long-horizon tasks with implicit preconditions. It features 2,628 expert demonstrations and 10,106 natural language task annotations across 57 scenes in the AI2-THOR 2.0 simulator \cite{kolve2017ai2}. The benchmark spans six task types which require agents to perform complex and temporally extended interactions such as picking, placing, cleaning, heating, cooling, and stacking. The distribution of these task types is detailed in Appendix~\ref{sec:construction_details}. Each annotations includes a task goal and step-by-step high-level descriptions, supporting hierarchical task understanding. By isolating dynamic affordance violations while keeping task structure and instruction format unchanged, DynAfford functions as a diagnostic benchmark that specifically evaluates an agent’s ability to infer and recover from unmet preconditions.

\subsection{Problem Statement}

We consider an instruction-conditioned embodied agent that executes high-level tasks in a dynamic household environment. Each episode begins with a natural language goal $G$ and a sequence of low-level instructions $L = \langle l_1, l_2, \ldots, l_n \rangle$, where $l_i$ denotes the $i$-th subgoal. The agent composes a sequence of skills from a predefined action library. At each time step $t$, it receives an RGB observation $o_t$ and instruction $l$, and selects the next action $a_t$. The task is challenging due to dynamic object affordances, where usability may change over time (e.g., a microwave being occupied) without being explicitly specified. The agent must infer implicit preconditions and reason about latent constraints. 
To evaluate this, we introduce a benchmark requiring agents to (1) interpret high-level goals, (2) track evolving object usability, and (3) adapt action sequences accordingly. Formally, the policy is defined as $\pi: (o_t, l) \mapsto a_t$ under partial observability and dynamic conditions.

\subsection{Dataset Construction}
\paragraph{Base Pipeline}
DynAfford is built upon the ALFRED trajectory generation pipeline, inheriting its task templates,
instruction annotations, and expert demonstration framework.
This design choice ensures direct comparability with prior embodied benchmarks,
while isolating dynamic object affordances as the primary source of distributional shift.

\paragraph{Affordance State Injection}
To introduce dynamic affordance violations, we intervene in the initial object states of each episode
by injecting affordance-specific constraints at task initialization.
For a given instruction $G$ and its associated low-level subgoals
$L = \langle l_1, l_2, \ldots, l_n \rangle$,
a subset of task-relevant objects is initialized in an \emph{Unavailable} state
according to the semantic preconditions implied by modifying object-level state attributes that encode semantic preconditions while keeping the instruction text and task goal unchanged.
We consider several categories of affordance unavailability, including:
(i) \textbf{Occupied objects}, where appliances such as microwaves are already in use;
(ii) \textbf{Used objects}, where containers (e.g., pans or plates) are unavailable due to prior usage;
and (iii) \textbf{Dirty objects}, where items such as cloths violate cleanliness-related preconditions.
During execution, agents must monitor object usability and adapt their action selection $a_t$ accordingly.

\paragraph{Static vs. Dynamic Split}
To enable controlled evaluation, DynAfford contains both \textbf{static} and \textbf{dynamic} affordance settings.
In static episodes, all objects satisfy their assumed preconditions, following the idealized setup of prior benchmarks.
In dynamic episodes, one or more target objects violate implicit affordance constraints at initialization,
requiring agents to detect and resolve unmet preconditions during execution.
Importantly, task goals, instruction text, and success criteria remain identical across the two settings,
ensuring that performance differences can be attributed solely to affordance reasoning. Roughly half of the dataset is constructed under each condition. Full task statistics and corresponding evaluation results are provided in Appendix~\ref{sec:task_distribution}.

\subsection{Data Splits}

Table~\ref{tab:dataset_distribution} presents the distribution of the dataset. The seen split includes 51 scenes, evenly divided between two room types: 27 kitchens and 24 bathrooms. The unseen split contains 6 scenes, with 3 from each room type, which are not included in the training data. This setup allows for evaluating both in-distribution generalization and performance under distribution shift.

Table~\ref{tab:obj_categories} provides a consolidated summary of the dataset,
including 2,076 unique test tasks (1,081 seen / 995 unseen)
and 7,351 goal conditions (3,969 seen / 3,382 unseen).
Unlike existing benchmarks, DynAfford covers a diverse set of object categories,
including appliances (e.g., microwaves), six types of tableware (mug, cup, bowl, plate, pot, pan),
cloth items, and complex multi-object scenarios.

Additional details on the construction of expert demonstrations, object affordance setting, and instruction annotation processes are provided in Appendix~\ref{sec:construction_details}.

\subsection{Design Rationale and Generalization}
DynAfford isolates dynamic affordance reasoning as a challenge orthogonal to perception noise. Even under perfect perception, execution-time availability shifts (e.g., occupancy or contamination) require agents to infer and resolve latent preconditions, making DynAfford a diagnostic benchmark for such capabilities.

Rather than introducing superficial perturbations, DynAfford modifies the task-generation pipeline by injecting dynamic state-transition rules that alter object availability during execution. 
We update the underlying FF planner~\cite{hoffmann2001ff} to regenerate expert demonstrations, retaining only tasks that remain fully solvable under the modified planning graph. 
Although built on ALFRED for comparability, the formulation is platform-agnostic, requiring only symbolic predicates and transition rules, and can be extended to other simulators or real-world systems.

DynAfford focuses on three representative affordance categories: \textit{Occupied}, \textit{Used}, and \textit{Dirty}, which capture common bottlenecks in long-horizon execution and require multi-step recovery. 
More complex affordances (e.g., geometric or structural constraints) are left for future extension but are naturally supported within this framework.

\begin{table}[t]
  \centering
  \small
  \setlength{\tabcolsep}{1mm}  
  \begin{tabular}{l c c c c c} 
    \toprule
    & \multirow{2}{*}{Train} 
    & \multicolumn{2}{c}{Validation} 
    & \multicolumn{2}{c}{Test} \\
    \cmidrule(lr){3-4} \cmidrule(lr){5-6}
    & & Seen & Unseen & Seen & Unseen \\
    \midrule
    \#Scenes         & 51    & 36   & 2   & 44   & 4 \\
    \#Demonstrations & 2628  & 189  & 217 & 228  & 223 \\
    \#Annotations    & 10106 & 772  & 872 & 1081 & 995 \\
    \bottomrule
  \end{tabular}
  \caption{DynAfford data splits. }
  \label{tab:dataset_distribution}
\end{table}


\begin{table}[t]
  \centering
  \footnotesize
  \setlength{\tabcolsep}{2pt}
  \begin{tabular}{l >{\centering\arraybackslash}m{2.5cm} c c c}
    \toprule
    Category & Available & Unavailable & Total \\
    \midrule
    Appliance & 86 / 249 & 128 / 157 & 214 / 406 \\
    Tableware & 82 / 204 & 605 / 316 & 687 / 520 \\
    Cloth & 46 / 49 & 106 / 0 & 152 / 49 \\
    Advanced & 0 / 0 & 28 / 20 & 28 / 20 \\
    \midrule
    Total & 214 / 502 & 867 / 493 & 1,081 / 995 \\
    \bottomrule
  \end{tabular}
  \caption{Task distribution of object categories and affordance conditions in DynAfford. Unavailability is introduced via structured state-transition rules, including occupancy conflicts, contamination, and precondition violations.
\textbf{Advanced} denotes tasks involving at least one appliance and one additional object from either tableware or cloth categories.}
  \label{tab:obj_categories}
\end{table}

\begin{figure*}[tb]
  \centering
  \includegraphics[width=1.0\linewidth]{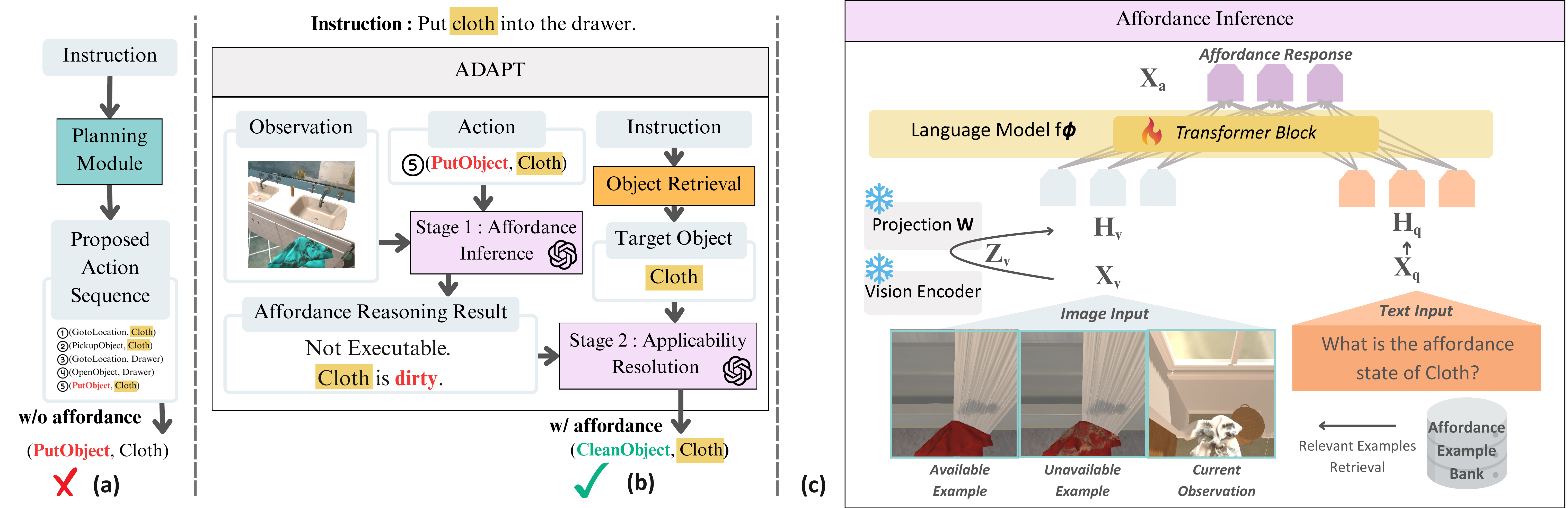}
  \caption{
  \textbf{Overview of ADAPT: Affordance-Driven Adaptive Planning and Task execution.}
  (a) Standard embodied instruction-following pipeline used in prior work, where the agent directly executes a planned action sequence without considering dynamic affordance constraints.
  (b) Our ADAPT framework augments action execution with affordance awareness.
For each proposed action, the agent first performs \emph{Stage 1: Affordance Inference} by jointly considering the current observation and the intended action.
If the action is considered inapplicable (e.g., the microwave is occupied), \emph{Stage 2: Applicability Resolution} selects an alternative executable action (e.g., waiting) instead of blindly executing the original plan.
(c) Architecture of the affordance inference module, which combines LoRA-finetuned visual grounding with multimodal in-context learning using retrieved affordance examples to infer object usability.
}
  \label{fig:AAS}
\end{figure*}

\section{Method}
Conventional embodied planning systems typically assume that actions are executable whenever they are linguistically valid, and handle failures by replanning from scratch, which can disrupt long-horizon coherence. 
In contrast, we model action execution as governed by latent, potentially time-varying preconditions.

\subsection{ADAPT: Affordance-Driven Adaptive Planning and Task execution}
We propose \textbf{ADAPT}, a decision-time module that enables agents to reason about action applicability under dynamic affordances. Given the current observation and a planned action, ADAPT determines executability and, if necessary, selects an alternative action that preserves task intent.

As shown in Figure~\ref{fig:AAS}, ADAPT operates through two steps: (1) \emph{affordance inference}, which evaluates whether preconditions are satisfied, and (2) \emph{applicability resolution}, which identifies an executable alternative when they are violated.

ADAPT is architecture-agnostic and integrates with existing agents without modifying planning or control components.

\paragraph{Stage I: Affordance State Inference}
ADAPT first infers whether the preconditions of the current high-level action are satisfied. Let $l_i \in L$ denote the planned high-level action, and let the \emph{target object} be the object explicitly referenced and manipulated by $l_i$. Affordance inference is triggered only when $l_i$ involves a target object known to exhibit dynamic affordance behavior (e.g., \textit{Microwave}, \textit{Cloth}).

\paragraph{LoRA Fine-Tuning}
To infer fine-grained affordance states (e.g., whether a cloth is clean or whether an appliance is available), we employ a vision-language model adapted via Low-Rank Adaptation (LoRA) \cite{hu2022lora}. Specifically, we fine-tune LLaVA-1.5-7B using training data constructed by replaying expert demonstrations from the DynAfford training split. Each example is labeled as \texttt{available} or \texttt{unavailable} based on task-specific latent preconditions. No data from the validation or test splits is used during fine-tuning.

\begin{figure}[tb]
  \centering
  \includegraphics[width=1.0\linewidth]{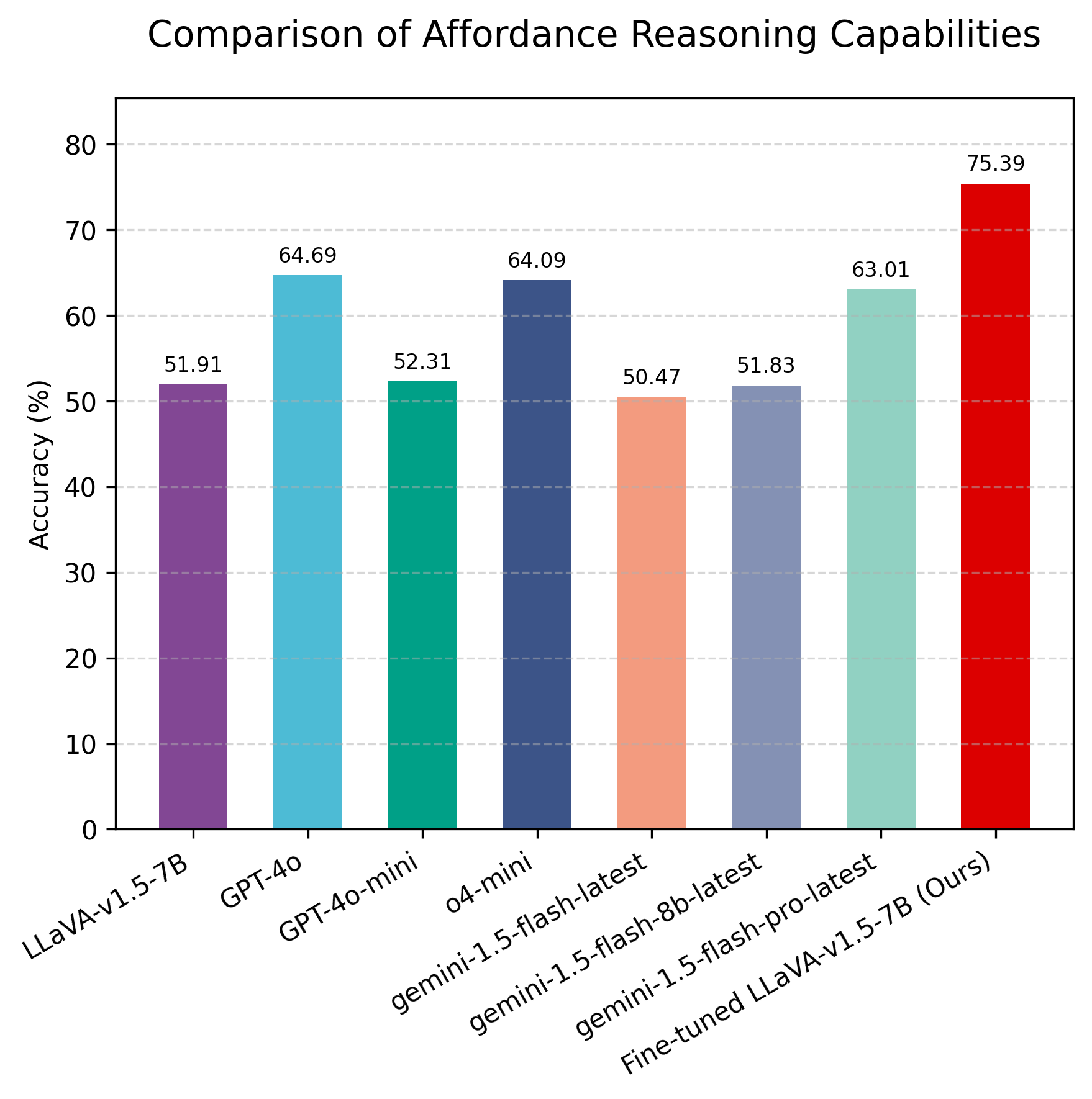}
  \caption{Isolates  the affordance inference component of ADAPT and compares it with several commercial vision-language models under the same input format.
Despite stronger general-purpose reasoning, commercial models underperform the LoRA-finetuned LLaVA on affordance inference by an average margin of more than 10\%.}
  \label{fig:ARC_overall}
\end{figure}

Figure~\ref{fig:ARC_overall} compares affordance prediction accuracy across multiple vision-language models. The fine-tuned model consistently outperforms general-purpose VLMs across object categories, demonstrating the effectiveness of task-specific adaptation for affordance grounding. This demonstrates the effectiveness of domain-specific fine-tuning for affordance understanding. A full breakdown of per-category results is provided in Appendix~\ref{sec:annotation_process}.

\paragraph{Multimodal In-Context Grounding}
To further strengthen affordance inference, ADAPT incorporates multimodal in-context grounding through templated inputs. Each query consists of three images: 
(1) a reference image depicting the target object in an available state, 
(2) a reference image depicting the object in an unavailable state, and 
(3) the current egocentric observation.

These images are concatenated and paired with a textual prompt describing the reference states and querying the usability of the current observation. Reference examples are retrieved from a held-out affordance example bank that does not overlap with training data. This structured multimodal context enables the model to reason about object usability by direct visual comparison, improving generalization under unseen configurations. Additional details on example retrieval and prompt construction are provided in Appendix~\ref{sec:affordance_reasoining}.

\paragraph{Stage II: Applicability Resolution}
When the inferred affordance state indicates that the current action is \texttt{unavailable}, ADAPT defers the execution of the action and temporarily suspends progress until the required condition is satisfied. Rather than replanning from scratch, ADAPT maintains commitment to the original intention and infers a resolution strategy that restores action applicability.

Resolution is performed by querying a large language model with a structured prompt encoding: 
(1) the current observation, 
(2) the inferred affordance constraint, and 
(3) the deferred high-level action. 
The model infers a commonsense-consistent resolution action, such as waiting for a temporary constraint to clear or executing a preparatory action (e.g., cleaning).

Once the constraint is resolved, ADAPT resumes execution of the originally deferred action. This mechanism preserves long-horizon task coherence while enabling flexible adaptation to dynamic state violations, without explicit error signals from the environment. Implementation details and prompt templates are provided in Appendix~\ref{sec:affordance_reasoner_details}.

\begin{table*}[t]
  \centering
  \setlength{\tabcolsep}{1mm}
  \begin{tabular}{l
                  S[table-format=2.2] S[table-format=2.2] 
                  S[table-format=2.2] S[table-format=2.2]
                  S[table-format=2.2] S[table-format=2.2]
                  S[table-format=2.2] S[table-format=2.2]}
    \toprule
    \multirow{2}{*}{\textbf{Method}} &
    \multicolumn{4}{c}{\textbf{Test Seen}} &
    \multicolumn{4}{c}{\textbf{Test Unseen}} \\
    \cmidrule(lr){2-5} \cmidrule(lr){6-9}
    & GC & PLW GC & SR & PLW SR & GC & PLW GC & SR & PLW SR \\
    \midrule
    \textbf{Few-Shot Methods} & & & & & & & & \\
    \midrule
    SayCan        & 4.79 & 3.58 & 0.46 & 0.11 & 10.50 & 6.71 & 0.00 & 0.00 \\
    LLM-Planner    & 7.16 & 1.84 & 1.11 & 0.54 & 15.34 & 4.33 & 2.46 & 1.33 \\
    \midrule
    \textbf{Supervised Methods} & & & & & & & & \\
    \midrule
    MOCA                 & 4.10 & 3.33 & 0.46 & 0.10 & 9.72 & 9.80 & 0.00 & 0.00 \\
    FILM             & 11.36 & 11.37 & 2.77 & 1.46 & 25.54 & 24.01 & 9.34 & 3.27 \\
    CAPEAM           & 20.10 & 14.44 & 9.25 & 4.56 & 36.28 & 31.39 & 19.39 & 7.87 \\
    \midrule
    \textbf{FILM + ADAPT (L)}           & 16.17 & 14.63 & 4.62 & 1.92 & 34.41 & 30.86 & 16.18 & 5.54 \\
    \textbf{CAPEAM  + ADAPT (L)}       & \textbf{22.95} & \textbf{19.29} & \textbf{10.82} & \textbf{7.28} & \textbf{37.43} & \textbf{37.45} & \textbf{21.10} & \textbf{10.94} \\
    \textbf{CAPEAM  + ADAPT (G)}       & 15.12 & 11.86 & 8.88 & 4.95 & 36.42 & 33.86 & 20.88 & 8.95 \\
    \bottomrule
  \end{tabular}
  \caption{
  Main results on the DynAfford benchmark. 
  \textbf{L} denotes LoRA-finetuned LLaVA, and \textbf{G} denotes GPT-4o.}
  \label{tab:main_results}
\end{table*}

\section{Experiments}
\subsection{Baselines}
To evaluate the generality of \textbf{ADAPT}, we integrate it into two supervised embodied agents, FILM \cite{min2021film} and CAPEAM \cite{kim2023context}. In all settings, the base planner, perception modules, and low-level controllers remain unchanged. ADAPT operates solely as a decision-time inference module that intercepts high-level actions involving dynamic objects and resolves their applicability.

\paragraph{Supervised Methods}
We compare FILM+ADAPT and CAPEAM+ADAPT against MOCA \cite{singh2021factorizing}, FILM \cite{min2021film}, and CAPEAM \cite{kim2023context}. To support deferred execution under temporary affordance violations, we extend the high-level action space with a \texttt{Wait} action and implement a corresponding low-level controller. For FILM, we additionally provide a BERT-based instruction encoder \cite{devlin2019bert} for compatibility. For CAPEAM, oracle subgoals and expert trajectories are provided to ensure that failures arise from missing affordance reasoning rather than subgoal generation.
In addition, we include a variant where the affordance inference backend of ADAPT is replaced with a commercial vision-language model (GPT-4o \cite{hurst2024gpt}), enabling a direct comparison between domain-adapted affordance modeling and general-purpose VLM reasoning.

\paragraph{Few-Shot Methods}
We evaluate LLM-Planner \cite{song2023llm} and SayCan \cite{ahn2022saycan}. For LLM-Planner, we compare the original prompting setup with a variant using demonstrations adapted to DynAfford’s dynamic affordance conditions. For SayCan, we provide a ground-truth visibility oracle to reduce the action space, granting it an advantage during both action feasibility scoring and affordance evaluation. 

All methods are evaluated on identical task instances from the DynAfford benchmark under both static and dynamic affordance settings. In static tasks, object usability remains fixed throughout the episode. In dynamic tasks, object affordances may be temporarily violated due to state changes or contextual constraints, requiring agents to infer unmet preconditions and adapt their actions. ADAPT is activated only under dynamic conditions, isolating its contribution to affordance-aware decision making.

\subsection{Evaluation Metrics}
We evaluate agent performance using four standard metrics for embodied instruction following. \textbf{Success Rate (SR)} measures whether the final task goal is achieved, while \textbf{Goal Condition Success Rate (GC)} reflects the proportion of goal predicates satisfied, allowing partial credit. To account for execution efficiency, we further report \textbf{Path-Length Weighted SR (PLW SR)} and \textbf{Path-Length Weighted GC (PLW GC)}, which weight outcomes by the ratio between the agent’s trajectory length and the expert demonstration length.

Notably, DynAfford adopts commonsense-driven goal conditions that encode implicit constraints on object usability, making GC a particularly informative metric for evaluating affordance-aware reasoning.

\subsection{ADAPT Consistently Improves Embodied Planning under Dynamic Affordances}
Table~\ref{tab:main_results} summarizes the main results. For few-shot methods, despite integrating with strong large language models, LLM-Planner \cite{song2023llm} and SayCan \cite{ahn2022saycan} struggle to handle dynamic object usability. Similarly, the supervised baseline MOCA \cite{singh2021factorizing} exhibits limited performance under dynamic affordance conditions, indicating that explicit affordance awareness is missing in existing approaches. As for other supervised methods, across both FILM and CAPEAM, integrating ADAPT leads to substantial and consistent performance improvements, demonstrating that ADAPT provides benefits across different planning architectures.

For FILM, the impact of ADAPT is particularly pronounced. On the test seen split, ADAPT improves success rate (SR) from 2.77 to 4.62 (+66.8\%) and goal condition completion (GC) from 11.36 to 16.17 (+42.3\%), accompanied by consistent gains in path-length weighted metrics. On the unseen split, the improvements are even larger: SR increases from 9.34 to 16.18 (+73.2\%), and GC improves from 25.54 to 34.41 (+34.7\%). These results indicate that ADAPT substantially enhances FILM’s robustness in both familiar and novel environments by enabling more reliable handling of temporarily inapplicable actions.

CAPEAM also benefits consistently from ADAPT, though with smaller relative gains due to its stronger baseline performance. On the seen split, CAPEAM+ADAPT improves SR from 9.25 to 10.82 (+17.0\%) and GC from 20.10 to 22.95 (+14.2\%), with notable improvements in path-length weighted metrics, reflecting more efficient execution. On the unseen split, SR increases from 19.39 to 21.10 (+8.8\%) and GC from 36.28 to 37.43 (+3.2\%), while PLW SR and PLW GC improve substantially (from 7.87 to 10.94 and from 31.39 to 37.45, respectively). These gains suggest that ADAPT helps CAPEAM better preserve long-horizon task coherence when affordance constraints arise, even when overall success rates are already high.

We further compare different vision-language backends within ADAPT by replacing the LoRA-finetuned LLaVA with a commercial large language model, GPT-4o \cite{hurst2024gpt}. While GPT-4o \cite{hurst2024gpt} provides strong general-purpose reasoning, it consistently underperforms the finetuned LLaVA within the ADAPT framework, particularly on seen environments. This result highlights the importance of task-specific visual grounding: effective affordance-aware action selection depends not only on reasoning capacity, but also on how well affordance-relevant visual cues are aligned with the task domain. Together, these findings demonstrate that ADAPT yields robust improvements across planners, and that domain-adapted affordance perception plays a critical role in its effectiveness.

All reported results are obtained from a single deterministic evaluation run on the full test split of DynAfford, with metrics aggregated over all episodes in each split. While commercial APIs may still exhibit minor nondeterminism, our evaluation focuses on affordance classification rather than open-ended generation, and the observed performance gaps are consistent across object categories.

\begin{table}[t]
  \centering
  \small
  \setlength{\tabcolsep}{3pt}
  \begin{tabular}{l c c}
    \toprule
    \multirow{2}{*}{\textbf{Method}} & 
    \multicolumn{2}{c}{\textbf{Test Unseen}}  \\
    \cmidrule(lr){2-3} 
    & SR & GC \\
    \midrule
    \multicolumn{3}{l}{\textit{Ablation Study}} \\
    \midrule
    Full method                        & 27.69 & 44.01 \\
    No LoRA fine-tuning                & 16.07 & 42.47 \\
    No Multimodal In-context Learning  & 6.15  & 33.49 \\
    \midrule
    \multicolumn{3}{l}{\textit{Comparison with Heuristic}} \\
    \midrule
    Capeam (Vanilla)         & 6.15  & 33.49 \\
    Capeam + ADAPT (Ours)      & 27.69 & 44.01 \\
    Capeam + Heuristic       & 46.15 & 52.99 \\
    \bottomrule
  \end{tabular}
  \caption{Ablation of ADAPT components and comparison with a symbolic heuristic baseline under dynamic affordance conditions.}
  \label{tab:ablation_study}
\end{table}

\begin{table}[t]
  \centering
  \small
  \setlength{\tabcolsep}{3pt}
  \begin{tabular}{l c c c c}
    \toprule
    Method & Total (s) & Init (s) & Infer (s) & Steps \\
    \midrule
    CAPEAM & 72.6 & - & - & 358 \\
    CAPEAM + ADAPT & 210.8 & 32.8 & 0.2 / call & 371 \\
    \bottomrule
  \end{tabular}
  \caption{Runtime comparison under dynamic affordance conditions.}
  \label{tab:runtime}
\end{table}

\subsection{Ablation Study}
We ablate the two grounding mechanisms that provide affordance evidence to ADAPT: (1) task-specific adaptation of the vision-language model via LoRA fine-tuning, and (2) multimodal in-context learning (MICL) using affordance exemplars. We do not ablate ADAPT into independent components, as action applicability inference is inherently joint.

As shown in Table~\ref{tab:ablation_study}, replacing the fine-tuned LLaVA with its pretrained counterpart reduces SR from 27.69\% to 16.07\%, highlighting the importance of task-specific adaptation. Further removing MICL, by excluding reference exemplars and using only the current observation, leads to a sharp drop in SR to 6.15\%. These results demonstrate that both LoRA fine-tuning and exemplar-guided multimodal prompting are critical for robust affordance grounding.

\paragraph{Comparison with Symbolic Heuristic.}
We compare ADAPT with a symbolic heuristic baseline that has access to ground-truth simulator states, representing an upper bound due to its reliance on privileged information.
Under dynamic affordance conditions, the heuristic achieves higher performance (46.15\% SR) than ADAPT (27.69\% SR), while both outperform the vanilla baseline (6.15\% SR).
This gap suggests that the primary challenge lies in reliable state inference from visual observations rather than planning.
ADAPT provides a practical trade-off by improving robustness without relying on privileged signals.

\subsection{Case Studies}
We present representative qualitative results in Appendix~\ref{sec:case_study}. In a dynamic task (\textit{``Microwave an egg and place it on the countertop''}), the baseline CAPEAM agent fails after 797 steps due to repeated execution of inapplicable actions. With ADAPT, the agent detects the temporary unavailability, waits, and resumes execution once conditions permit, completing the task in 206 steps.

We further identify several recurring failure modes under dynamic affordance conditions:
(i) \textit{latent state mis-detection}, where the agent fails to recognize execution-time state changes;
(ii) \textit{cascading precondition violations}, where unresolved mismatches propagate through subsequent steps;
and (iii) \textit{incorrect recovery strategies}, where the agent performs redundant or suboptimal actions.

A representative failure case under static conditions is shown in Appendix~\ref{sec:failure_case}, where ADAPT misclassifies a partially visible object due to occlusion. While rare, such cases highlight limitations in visual grounding and motivate future work on viewpoint selection and object disambiguation.

\subsection{Efficiency and Runtime Analysis}

We measure wall-clock execution time on an NVIDIA RTX 3090.
As shown in Table~\ref{tab:runtime}, incorporating ADAPT increases total task duration (72.6s $\rightarrow$ 210.8s) compared to the baseline.
However, this increase is primarily due to longer interaction sequences required by dynamic affordance conditions
(e.g., waiting for appliances or resolving object states), rather than computational overhead.

Importantly, ADAPT operates in an event-driven manner rather than at every step.
The vision-language model is invoked only when affordance ambiguity arises,
resulting in a low average inference cost of approximately 0.2s per call.
Similarly, affordance resolution incurs a comparable cost.
The one-time initialization overhead (32.8s) does not scale with task length.

Overall, these results indicate that the additional computation introduced by ADAPT is modest,
and that runtime is dominated by environment interactions rather than model inference,
demonstrating the practical feasibility of the approach.

\section{Conclusion}
We introduce \textbf{DynAfford}, a new benchmark for evaluating embodied agents under dynamic object affordances and commonsense-driven constraints. Unlike prior benchmarks built under idealized assumptions, DynAfford explicitly requires agents to reason about latent and evolving preconditions that govern action applicability.

To address these challenges, we propose \textbf{ADAPT}, a unified decision-time inference mechanism that enables agents to assess object usability and defer execution of temporarily inapplicable actions. By treating affordance as a latent precondition rather than a static property, ADAPT allows agents to preserve long-horizon task coherence while adapting to dynamic environmental constraints.

When integrated into strong embodied agents such as FILM and CAPEAM, ADAPT consistently improves both task success rate (SR) and goal condition completion (GC), with the largest gains observed under dynamic affordance settings. On the test unseen split, ADAPT yields substantial improvements on FILM and consistent gains on CAPEAM, demonstrating its effectiveness in recovering from unmet preconditions and reducing brittle instruction-following behavior.

Together, DynAfford and ADAPT highlight a critical yet underexplored aspect of embodied intelligence: the ability to reason not only about \emph{what} action to take, but also \emph{when an action should not be executed}. We hope this work encourages future research on affordance-aware decision making and robust execution under latent and evolving constraints.

\section*{Limitations}
This work focuses on affordance reasoning under single-view egocentric observations to ensure comparability with existing embodied benchmarks and methods. As a result, ADAPT may be sensitive to partial occlusion and viewpoint ambiguity in certain cases, which can lead to incorrect affordance inference. Future work could incorporate multi-view perception or active viewpoint selection to improve robustness under visually challenging conditions. Finally, expanding DynAfford to capture richer physical variability and real-world interaction patterns remains an important direction for future work.

\section*{Potential Risks}
While DynAfford and ADAPT are designed as diagnostic tools for studying affordance-aware decision making in simulated household environments, they do not model real-world physical uncertainty, safety constraints, or human–robot interaction dynamics. Consequently, direct deployment without additional safeguards could lead to inappropriate action deferral or overly conservative behavior in real settings. We emphasize that our framework is intended for research evaluation rather than immediate real-world use, and extending it to physical robots would require integrating safety-aware control, uncertainty modeling, and human-in-the-loop supervision. 

\section*{Acknowledgements}
This work was supported in part by National Science and Technology Council, Taiwan, under Grant NSTC 112-2221-E-002-131-MY3.

\bibliography{custom}

\appendix

\section{Embodied AI Benchmarks: A Comparative Perspective}    
\label{sec:comparative_perspective}
Unlike most prior benchmarks that lack explicit goal conditions or dynamic affordance modeling, DynAfford uniquely integrates \textit{commonsense-driven goal conditions}, supports \textit{dynamic object affordances}, and uses \textit{natural language instructions} for \textit{long-horizon tasks}. While benchmarks such as BEHAVIOR-1K \cite{li2023behavior} and ALFRED \cite{shridhar2020alfred} support complex tasks, their goal representations are symbolic or static, and do not capture context-sensitive object usability. A summary comparison is provided in Table~\ref{tab:dataset_comparison}.

\section{Dataset Construction Details}
\label{sec:construction_details}
\subsection{Expert Demonstrations}
\begin{figure}[tb]
  \centering
  \includegraphics[width=1.0\linewidth]{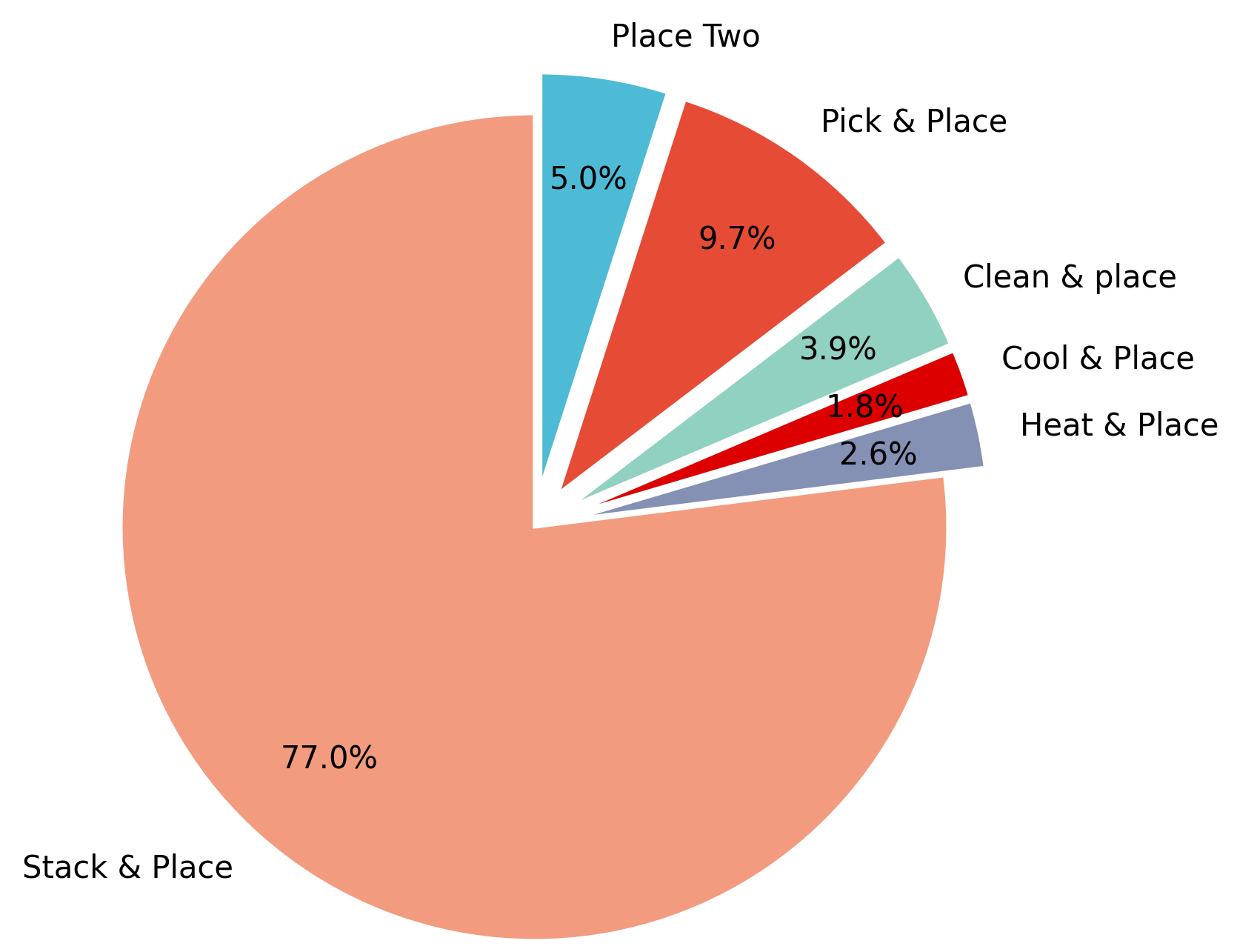}
  \caption{\textbf{Task types and annotation statistics in DynAfford.}
DynAfford contains over 1,000 demonstrations across 6 task types of varying complexity, each paired with 3–6 language instructions. Tasks range from simple placement to hierarchical stacking involving movable and fixed containers.
}
  \label{fig:task_type}
\end{figure}

We generate expert demonstrations by extending the ALFRED augmentation pipeline, and define six distinct task types in DynAfford which require agents to perform complex and temporally extended interactions such as picking, placing, cleaning, heating, cooling, and stacking. The distribution of these task types is illustrated in Figure~\ref{fig:task_type}. For each task type, we modify the PDDL domain to include dynamic-state predicates, For example, we add the predicates \texttt{(cleanable ?mo)} and \texttt{(isClean ?mo)} to model the “Dirty/Clean” scenario. This addition enables the Fast-Forward planner to interleave the necessary cleaning actions with the primary task, as illustrated in Listing~\ref{lst:pddl-domain}.

\subsection{Object Affordance Setting}
To simulate real-world variability, object states in DynAfford are randomized at the start of each task. Objects are labeled as either \emph{available} or \emph{unavailable}, with semantics depending on the object type. For example, a microwave is unavailable when occupied, and a cloth is unavailable when dirty.

\subsection{Task Complexity}
To control task complexity, we define two difficulty levels. 
\textbf{Basic} episodes contain at most one unavailable object, allowing agents to solve the task with minimal adaptation. 
In contrast, \textbf{Advanced} episodes include up to two unavailable objects, which require more complex reasoning, object monitoring, and contingency planning to complete the goal.

\subsection{Annotation Process}
\label{sec:annotation_process}
Traditional embodied AI benchmarks generate natural language instructions through a two-step process: first, expert demonstrations are synthesized using a deterministic planner, and then crowd workers on platforms such as Mechanical Turk manually write instructions based on the demonstration videos. While this approach yields human-readable annotations, it is time-consuming and often inconsistent across annotators.

To address these limitations and achieve high-quality yet efficient data generation, we adopt a semi-automated annotation pipeline composed of four stages: (1) \textbf{Template Refinement}, applying goal templates to improve completeness; (2) \textbf{LLM Paraphrasing}, using prompt-based language models to diversify phrasing; (3) \textbf{OOV Detection}, replacing out-of-vocabulary or inconsistent terms to improve linguistic reliability; and (4) \textbf{Human Verification}, where trained annotators perform lightweight review and correction to ensure clarity and task feasibility.

\section{Task Distribution and Evaluation Results}
\label{sec:task_distribution}
To ensure fair and comprehensive evaluation, DynAfford includes both \textbf{static} and \textbf{dynamic} object affordance settings.
\subsection{Static object affordance tasks} follow an idealized setting where object usability remains constant throughout the episode. These tasks simulate simplified environments commonly used in prior work and serve as a controlled benchmark to evaluate whether models perform comparably to existing methods under traditional assumptions. Table~\ref{tab:static_tasks_distribution} presents the distribution of static object affordance tasks, and corresponding results are reported in Table~\ref{tab:static_results}.
\subsection{Dynamic object affordance tasks} involve episodes where object usability may change during execution. These tasks test an agent's ability to detect shifting preconditions, monitor object usability, and recompose the policy plan as necessary. Table~\ref{tab:dynamic_tasks_distribution} presents the distribution of dynamic object affordance tasks, and corresponding results are summarized in Table~\ref{tab:dynamic_results}.

Approximately half of the tasks in DynAfford are static and the other half are dynamic, enabling fine-grained comparison between models under both simplified and realistic conditions. Among the static tasks, a subset is directly reused from the original ALFRED \cite{shridhar2020alfred} benchmark, ensuring compatibility and grounding in previously validated scenarios.

\section{Affordance Inference Capability}
\label{sec:affordance_reasoining}
\begin{figure*}[tb]
  \centering
  \includegraphics[width=1.0\linewidth]{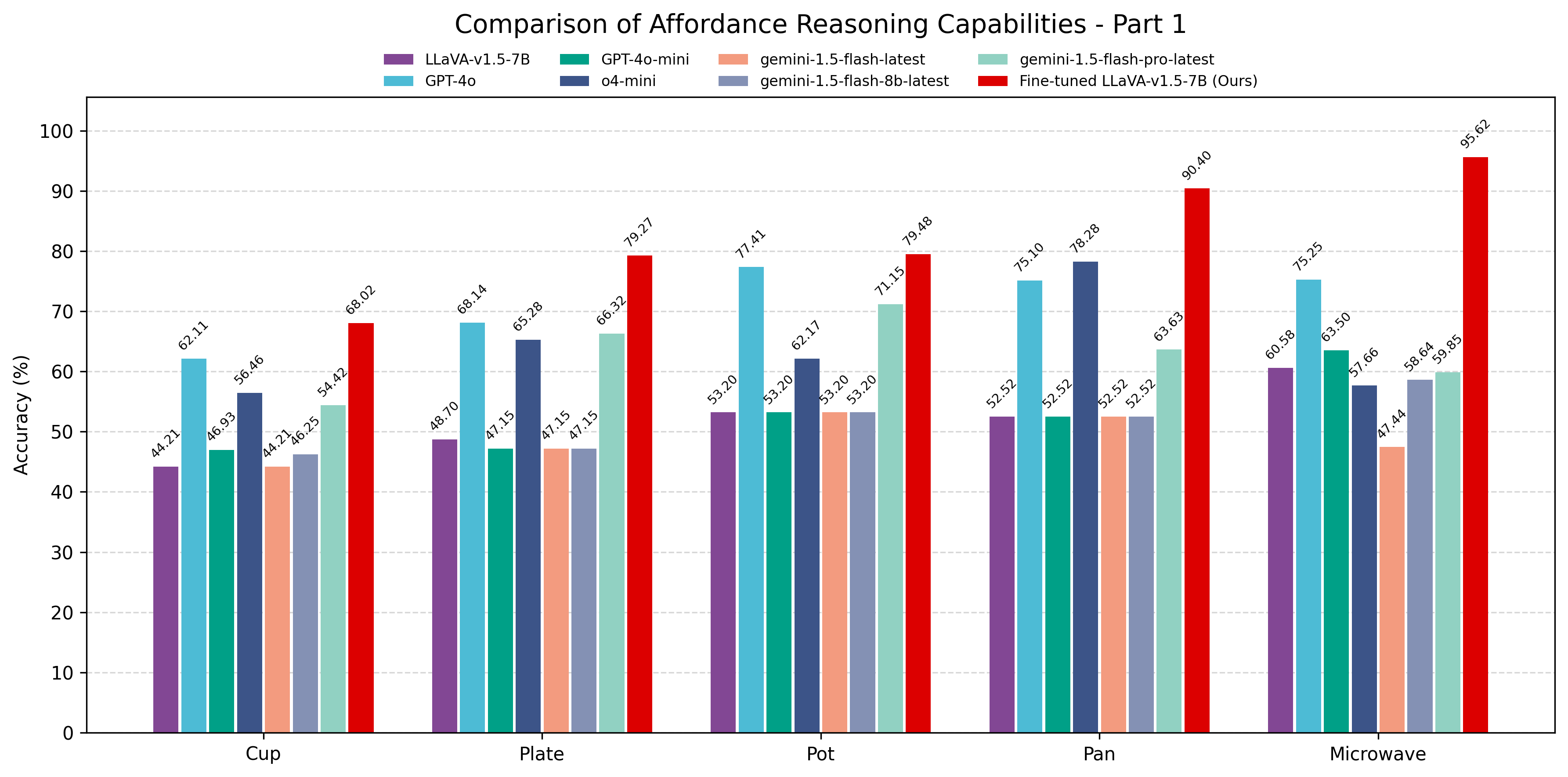}
  \caption{
  \textbf{Object Affordance Prediction Accuracy on Visually Distinguished Objects.}
  Our fine-tuned LLaVA-v1.5-7B model outperforms all baselines on large objects (e.g., Pot, Pan, Microwave) where visual cues are more spatially distinguished in the observation space. These results highlight the model’s ability to leverage strong visual evidence for dynamic affordance reasoning.
  }
  \label{fig:ARC_1}
\end{figure*}

Figure~\ref{fig:ARC_1} presents affordance prediction accuracy on five major object categories: Cup, Plate, Pot, Pan, and Microwave. Our fine-tuned LLaVA-v1.5-7B achieves the highest performance across all five categories, with substantial margins over both the base LLaVA and state-of-the-art general-purpose models. Notably, the model attains 90.40\% accuracy on Pan and 95.62\% on Microwave, demonstrating its strong capacity to reason about container-related affordances.

\begin{figure*}[tb]
  \centering
  \includegraphics[width=1.0\linewidth]{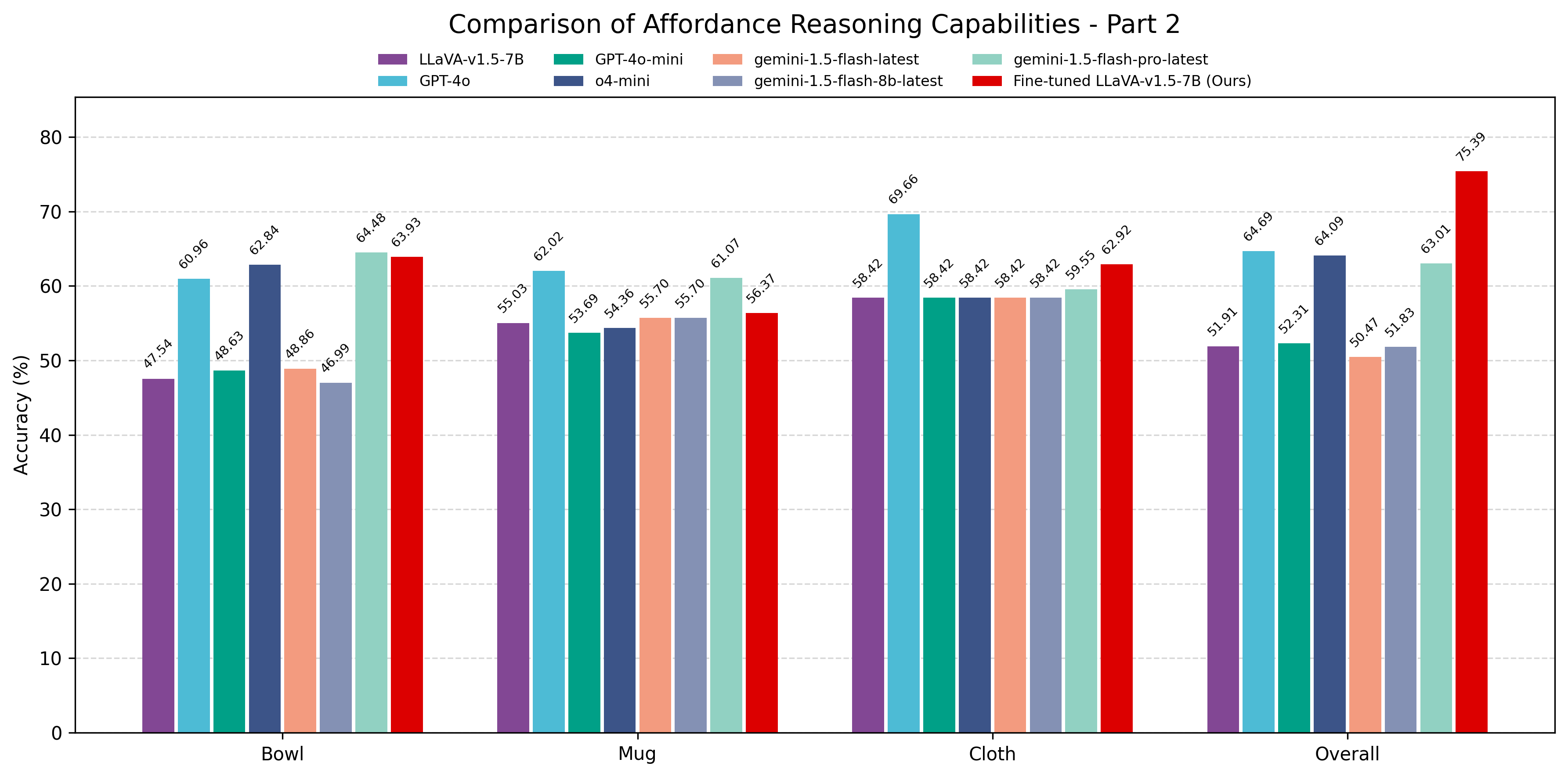}
  \caption{
  \textbf{Object Affordance Prediction Accuracy on Other Objects and Overall Performance.}
  While the gains from fine-tuning are less pronounced on smaller or more deformable objects (e.g., Mug, Cloth), our model still achieves competitive performance compared to much larger models like GPT-4o and Gemini-Pro. Overall accuracy reaches 75.39\%, validating the effectiveness of task-specific adaptation.
  }
  \label{fig:ARC_2}
\end{figure*}

We hypothesize that this performance advantage arises in part from the distinguished visual presence of these objects in the agent’s field of view. Objects like Pot, Pan, and Microwave typically occupy a large portion of the observation, providing more distinct spatial and contextual cues that facilitate affordance grounding. In contrast, smaller or more deformable objects, such as cloths or mugs, may present more subtle affordance shifts, which are further discussed in Figure~\ref{fig:ARC_2}. Nevertheless, when aggregated across all object categories, our fine-tuned model still outperforms all state-of-the-art general-purpose models, demonstrating its robustness and generalizability in affordance reasoning.

These results highlight the fine-tuned model’s ability to combine visual grounding with commonsense reasoning, enabling it to consistently outperform both lightweight and large-scale foundation models on affordance-sensitive categories.

\textbf{The code and partial dataset used in this evaluation are released as part of this submission; please refer to Appendix~\ref{sec:code_data} for details.}

\section{Affordance Reasoner Implementation}
\label{sec:affordance_reasoner_details}
\subsection{Visibility Detection}
To determine whether the target object is currently visible to the agent, we use a pretrained LLaVA model. The model is queried only if the current high-level action involves an affordance-critical object and goal visibility is uncertain.

\begin{figure*}[tb]
  \centering
  \includegraphics[width=1.0\linewidth]{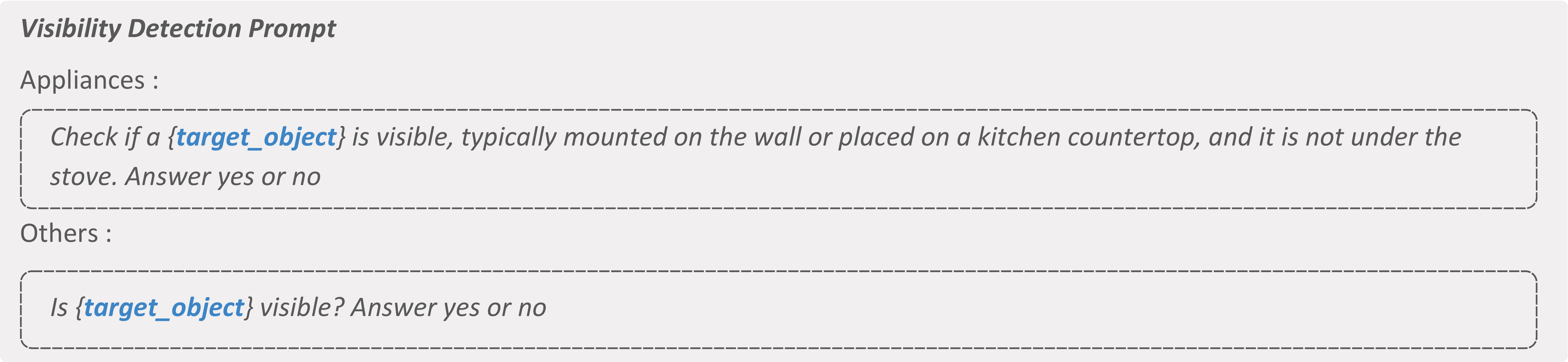}
  \caption{\textbf{Prompt format for visibility detection.}
We use an pretrained LLaVA-1.5-7B model to assess whether the target object is visible in the current egocentric view. The prompt is dynamically adapted based on the target object’s category, guiding the model to make accurate visibility judgments.}
  \label{fig:visibility_detection_prompt}
\end{figure*}

During inference, LLaVA receives the full egocentric frame as input. The accompanying prompt is dynamically adapted according to the object type, as detailed in Figure~\ref{fig:visibility_detection_prompt}.

If the response indicates the object is visible, the process proceeds to the next step.

\subsection{Affordance Detection}
\begin{figure*}[tb]
  \centering
  \includegraphics[width=1.0\linewidth]{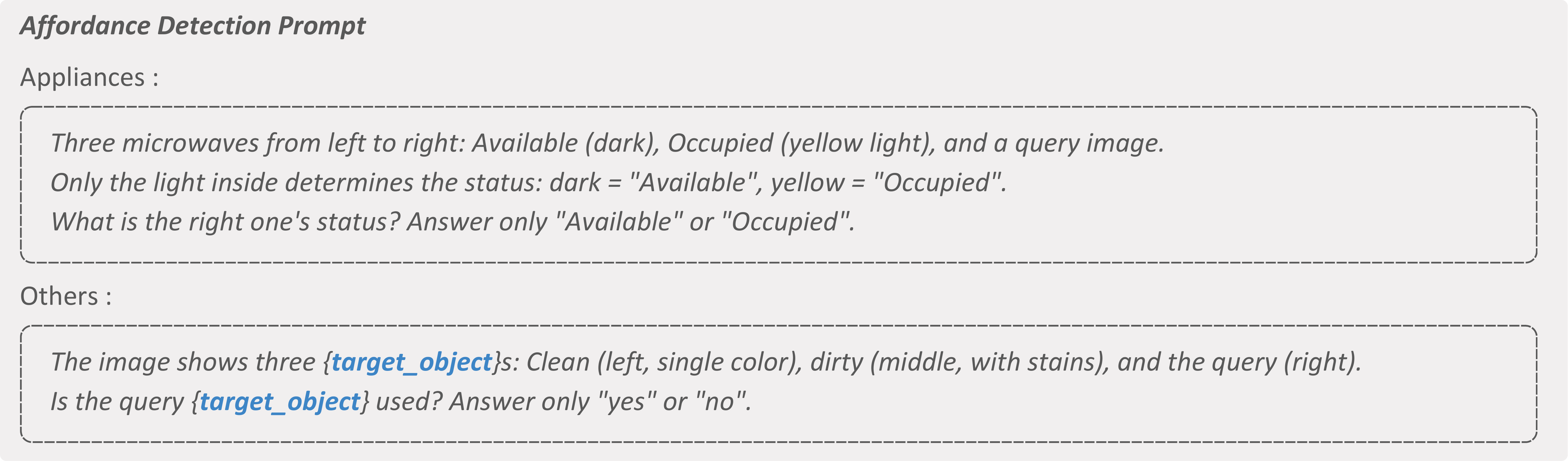}
  \caption{
  \textbf{Prompt format for affordance detection.}
We query the fine-tuned model with structured prompts that include visual and textual context to determine the current usability of a target object.}
  \label{fig:affordance_detection_prompt}
\end{figure*}

We empirically found that the pretrained LLaVA model performed poorly in detecting fine-grained usability attributes, such as whether a cloth is clean or a microwave is currently in use. To address this limitation, we fine-tuned LLaVA-1.5B using LoRA on a dataset collected by replaying expert demonstrations from ALFRED. This fine-tuning dataset is entirely separate from the DynAfford test split, ensuring a fair evaluation of affordance state recognition, as illustrated in Figure~\ref{fig:affordance_detection_prompt}. 

\begin{figure*}[tb]
  \centering
  \includegraphics[width=1.0\linewidth]{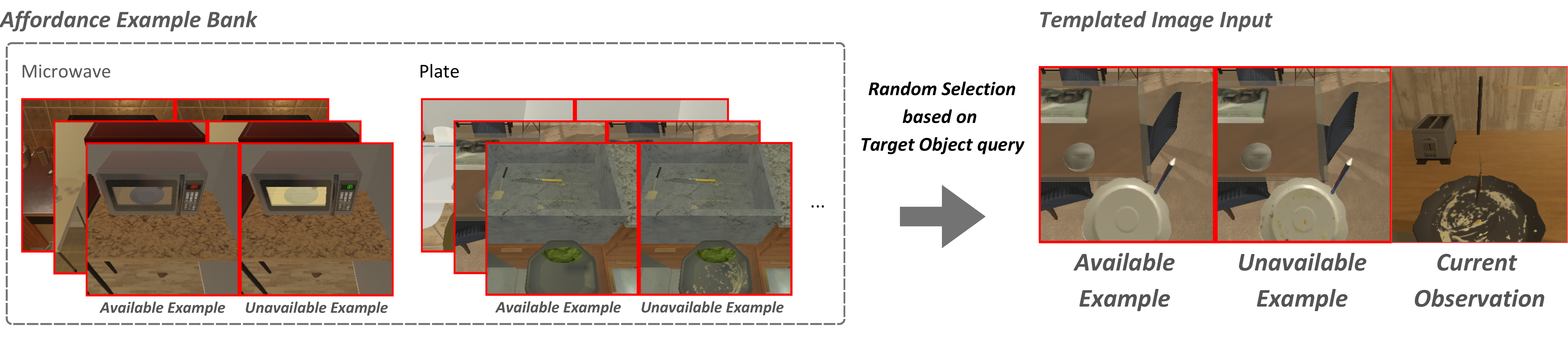}
  \caption{\textbf{Templated input for multimodal affordance reasoning.} Each input consists of a triplet of images—(1) a reference of the object in an available state, (2) a reference in an unavailable state, and (3) the current egocentric frame—paired with a textual prompt to assess the object’s current usability.}
  \label{fig:multimodal_ICL}
\end{figure*}

To further enhance prediction performance, we incorporate \textbf{multimodal in-context learning} (ICL) via templated input construction. As illustrated in Figure~\ref{fig:multimodal_ICL}, each input to the model consists of a triplet of images: (1) a reference image showing the object in an available state, (2) a reference in an unavailable state, and (3) the current egocentric frame. These are concatenated and paired with a textual prompt describing the reference images and querying the usability of the current frame.

At inference time, visual examples are retrieved from a held-out affordance example bank that does not overlap with fine-tuning data. This structured prompting improves generalization and enables the model to reason about object usability by comparing the current observation with contextualized examples. 

\label{sec:action_aware_memory_details}
\subsection{High-level Action Replanning}
\begin{figure*}[tb]
  \centering
  \includegraphics[width=1.0\linewidth]{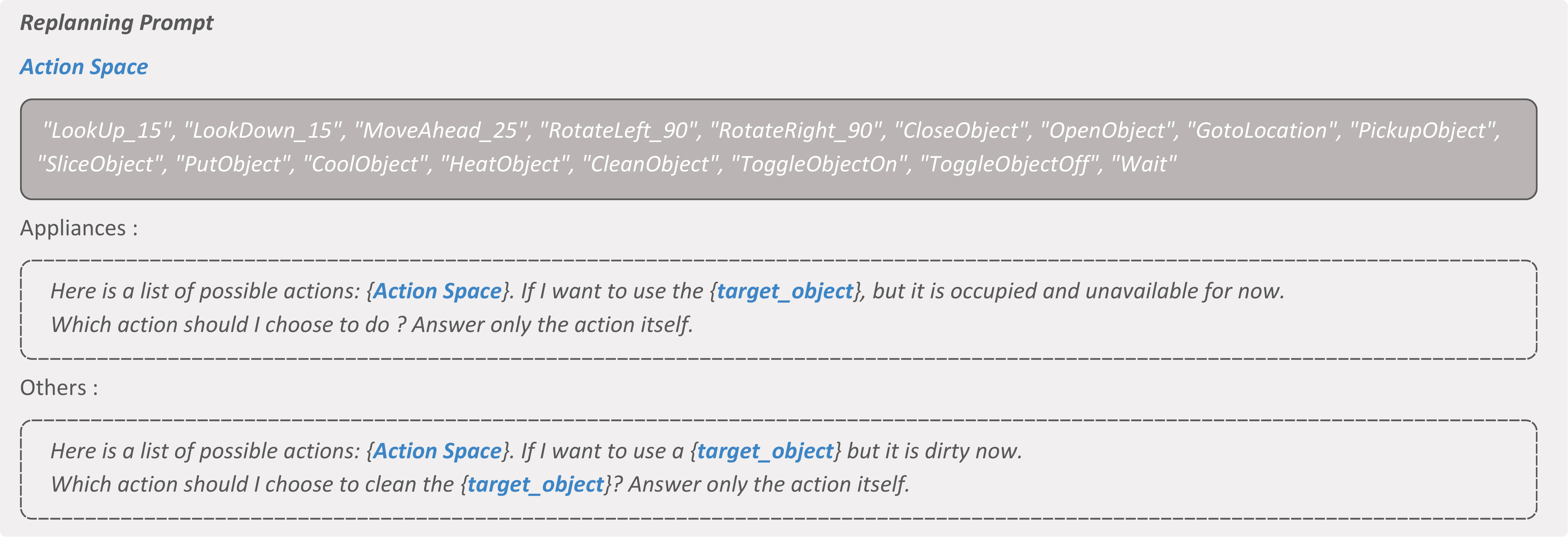}
  \caption{
  \textbf{Prompt format for high-level action replanning.}
The large language model receives contextual prompts containing the agent's current observation, high-level action list, and affordance status to infer the next appropriate action.}
  \label{fig:replanning_prompt}
\end{figure*}

When the \textbf{Affordance Inference Stage} determines that the preconditions of a high-level action are violated, e.g., when the target object is in an \textbf{Unavailable} state, the system performs applicability resolution via LLM-based inference. A contextualized prompt is constructed using the current observation, the target object and its affordance status, and the set of available high-level actions. Based on this information, the LLM infers an alternative executable action to resolve the unmet condition, and the agent updates its task plan accordingly by inserting or reordering subgoals. An example prompt is shown in Figure~\ref{fig:replanning_prompt}.

The inferred resolution strategy depends on the nature of the affordance violation. For temporary constraints (e.g., an occupied appliance or blocked interaction), the agent executes a \texttt{Wait} action and periodically re-evaluates the affordance status. For persistent constraints (e.g., a dirty object), the agent inserts a corrective subgoal such as cleaning, temporarily placing any carried items on a nearby surface before resuming the original task.

\section{Case Study}
\label{sec:case_study}
\begin{figure*}[tb]
  \centering
  \includegraphics[width=0.9\linewidth]{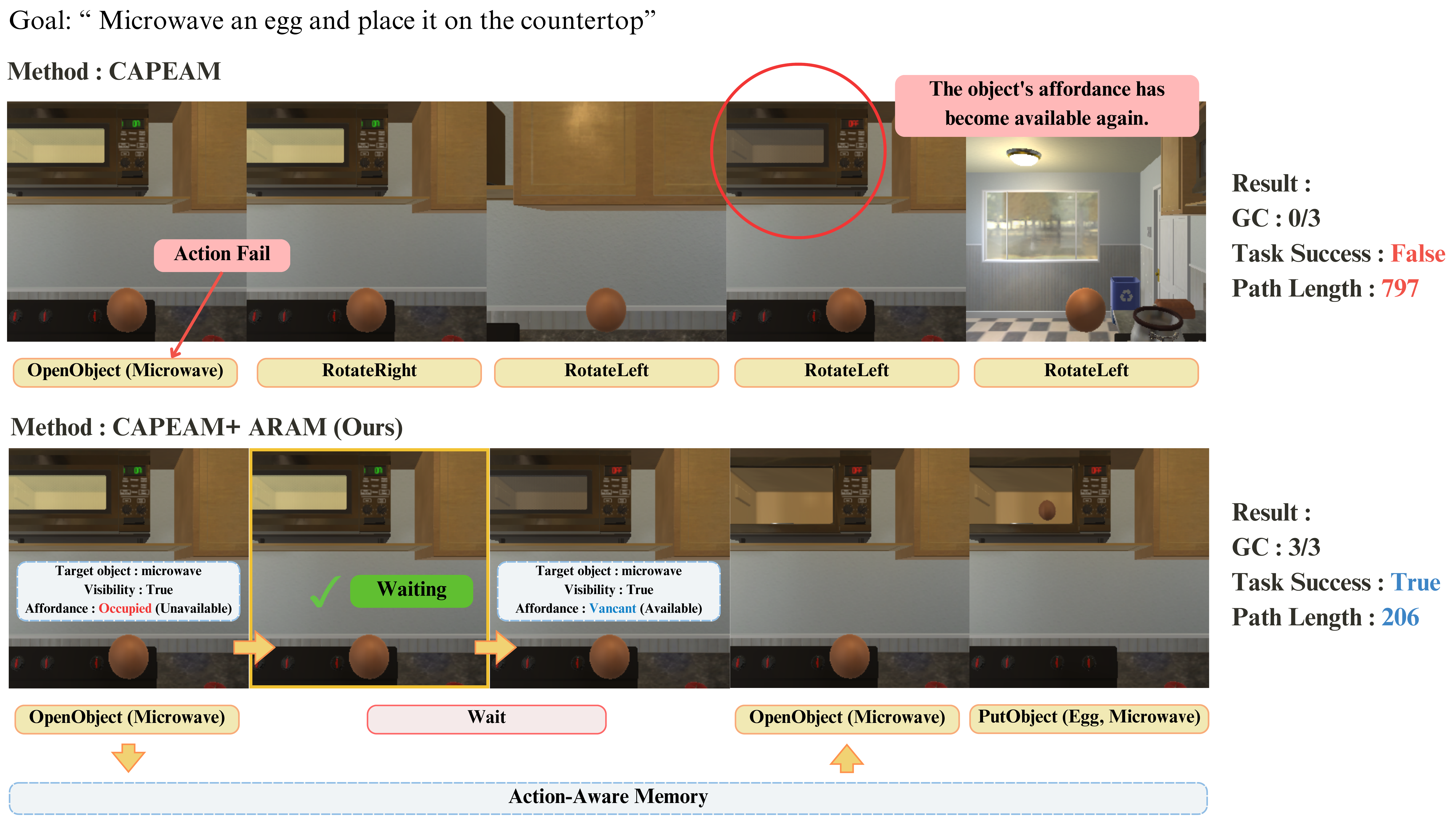}
  \caption{
  \textbf{ARAM improves robustness in dynamic environments.} 
 In the task “Microwave an egg and place it on the countertop,” CAPEAM fails due to an occupied microwave and takes 797 steps. With ARAM, the agent detects the temporary unavailability, waits, and resumes the pending action, completing the task in only 206 steps—demonstrating improved adaptability in dynamic settings.
}
  \label{fig:case_study}
\end{figure*}

Figure~\ref{fig:case_study} shows a dynamic object affordance task: \textit{"Microwave an egg and place it on the countertop."} In this scenario, the baseline CAPEAM \cite{kim2023context} agent navigates to the microwave and attempts to open it. However, the microwave is initially in an \texttt{Unavailable} state, rendering it temporarily unusable. This triggers a failed action, after which the model enters a replanning loop and repeatedly predicts a \texttt{RotateRight} action followed by a \texttt{RotateLeft}, ultimately returning to its original position.

Although the microwave becomes available again during this loop, CAPEAM fails to resume the prior goal of opening it, as it lacks memory of the failed action. The agent continues rotating aimlessly until the episode ends unsuccessfully after 797 steps.

In contrast, the ARAM-enhanced CAPEAM first detects the \texttt{Unavailable} state of the microwave using the affordance reasoner and stores the \texttt{OpenObject} action as a pending action in the Action-Aware Memory (AAM). It then executes a \texttt{Wait} action and periodically reassesses the microwave's affordance. Once the microwave becomes available, the pending action is retrieved and executed. The agent proceeds to complete the task successfully in just 206 steps. This case demonstrates ARAM’s ability to support flexible recovery and efficient replanning under dynamic state changes.

\section{Failure Case}
\label{sec:failure_case}

\begin{figure*}[tb]
  \centering
  \includegraphics[width=0.9\linewidth]{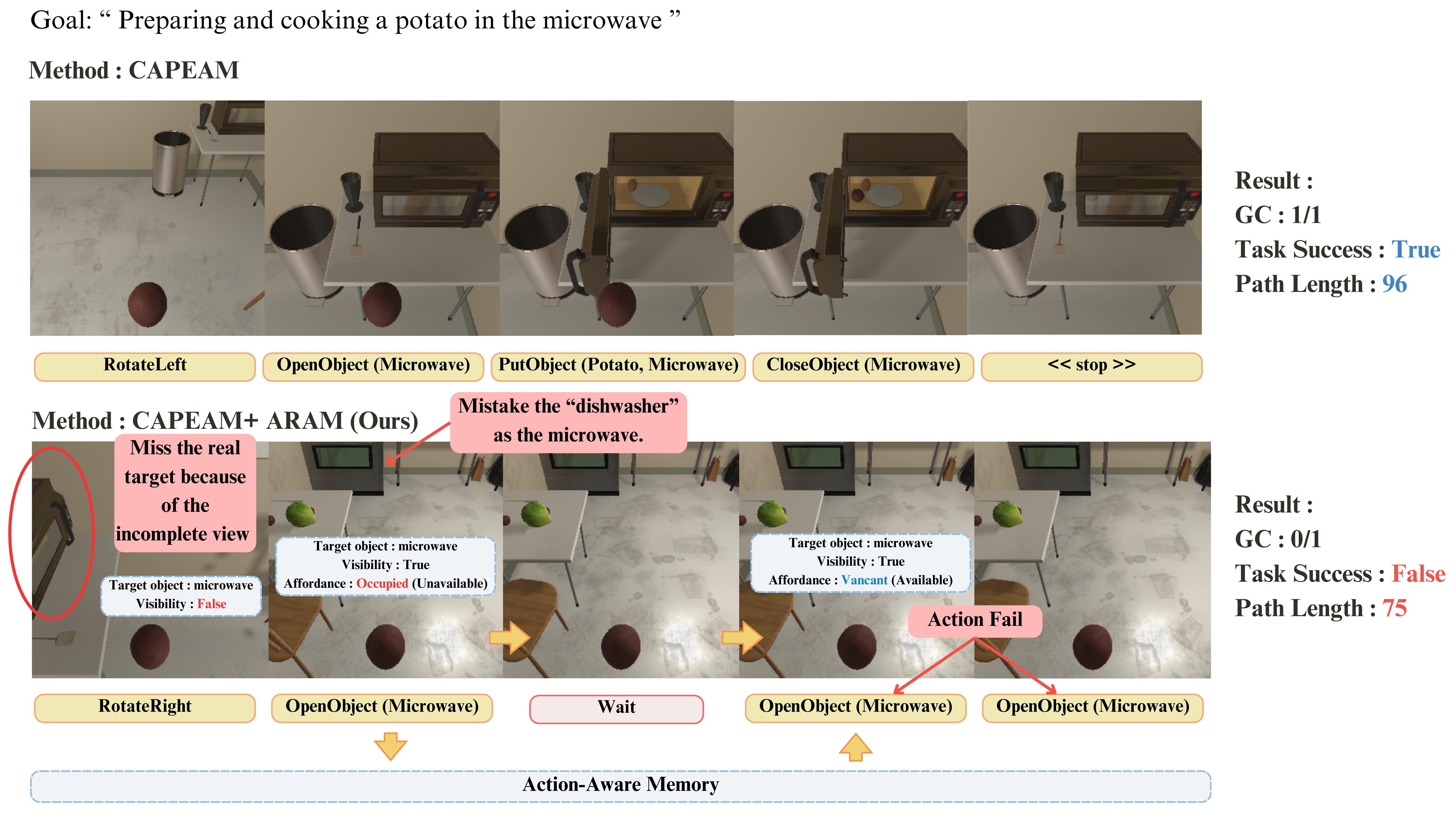}
  \caption{
  \textbf{Failure case due to misidentification.}
  In this failure case, the agent misclassifies a dishwasher as the microwave due to partial occlusion, triggering incorrect affordance reasoning and ultimately leading to task failure.}
  \label{fig:failure_case}
\end{figure*}

Figure~\ref{fig:failure_case} presents a failure case on a static affordance task: \textit{"Prepare and cook a potato in the microwave."} In this example, the baseline CAPEAM agent behaves as expected by navigating to the microwave, placing the potato inside, and successfully completing the task.

However, when ARAM is integrated, the affordance reasoner encounters partial occlusion: the microwave is only half visible from the agent’s current pose. As a result, the model incorrectly classifies the microwave as \texttt{Invisible}. The agent turns away and mistakenly identifies a nearby dishwasher as the target object. It then incorrectly predicts the dishwasher’s affordance as \texttt{Unavailable} and begins issuing \texttt{Wait} actions. After a while, the affordance flips to \texttt{Available}, and the model attempts to execute an \texttt{OpenObject} action on the dishwasher, leading to a failed trajectory.

This failure results from compounded errors in object perception and affordance reasoning. While such cases are rare, they highlight limitations in affordance reasoning under occlusion and visually ambiguous contexts. Addressing these challenges, such as through improved viewpoint selection or enhanced object disambiguation, remains an important direction for future work.

\section{Use Of AI Assistants}
\label{sec:ai_usage}
During the course of this work, we made limited use of AI-assisted tools as auxiliary aids.
These tools were used primarily to improve the presentation quality of the manuscript,
including enhancing readability, refining phrasing, and assisting with minor code refactoring.
Importantly, the conception of the research problem, benchmark design,
methodological development, and experimental evaluation were entirely carried out by the authors.
AI-assisted tools did not contribute to model design, data construction,
or scientific decision-making.

\section{Code and Data Availability}
\label{sec:code_data}
We provide code and data to reproduce the main experiments in our code appendix. The released package includes:
\begin{itemize}
    \item \textbf{Benchmark Evaluation Data:} 
    Test split data (both seen and unseen) of the DynAfford benchmark
    
    \item \textbf{Affordance Reasoning Evaluation:}
    We provide a subset of validation data for evaluating affordance reasoning with our LoRA fine-tuned LLaVA-1.5-7B and other vision-language models, including multimodal in-context prompt logs and the corresponding visual assets used during inference.
    
    \item \textbf{Preliminary Evaluation Results:}
    JSON files containing prediction outputs from both fine-tuned and pretrained models over the full DynAfford test set (1,252 samples)

    \item \textbf{Codebase:}
    Scripts for dataset statistics, affordance prediction evaluation. A local copy of the LLaVA repository is included for convenience.
\end{itemize}


\section{Fine-Tuning and Computing Infrastructure Details}
\label{sec:implementation}

We fine-tuned the LLaVA v1.5-7B model using LoRA, built on top of the Vicuna-7B base language model and the CLIP ViT-L/14-336 vision encoder. Training was performed on a single NVIDIA RTX 3090 GPU (24GB VRAM) using CUDA 11.8 and PyTorch 2.6.0. The model was trained for one epoch with a LoRA rank of 64, a learning rate of 1e-5, and a batch size of 4 per device.

\vspace{1em}
\noindent
\begin{minipage}{\linewidth}
\centering
\small
\setlength{\tabcolsep}{3pt}
\vspace{0.1em}
\begin{tabular}{l c c c c c} 
  \toprule
  \textbf{} 
    & \multirow{2}{*}{\textbf{Train}} 
    & \multicolumn{2}{c}{\textbf{Validation}} 
    & \multicolumn{2}{c}{\textbf{Test}} \\
  \cmidrule(lr){3-4} \cmidrule(lr){5-6}
    & 
    & \textbf{Seen} & \textbf{Unseen}
    & \textbf{Seen} & \textbf{Unseen} \\
  \midrule
  \#Scenes            & 51              & 36               & 2              & 44            & 4              \\
  \#Demonstrations    & 1580  & 98               & 154           & 46             & 122         \\
  \#Annotations       & 5415  & 360                & 575           & 214            & 502              \\
  \bottomrule
\end{tabular}
\captionof{table}{\textbf{Static Object Affordance Task Split} 
This table summarizes the subset of DynAfford tasks with static object usability throughout the episode.}
\label{tab:static_tasks_distribution}
\end{minipage}

\vspace{1em}
\noindent
\begin{minipage}{\linewidth}
\centering
\small
\setlength{\tabcolsep}{3pt}
\vspace{0.1em}
\begin{tabular}{l c c c c c} 
  \toprule
  \textbf{} 
    & \multirow{2}{*}{\textbf{Train}} 
    & \multicolumn{2}{c}{\textbf{Validation}} 
    & \multicolumn{2}{c}{\textbf{Test}} \\
  \cmidrule(lr){3-4} \cmidrule(lr){5-6}
    & 
    & \textbf{Seen} & \textbf{Unseen}
    & \textbf{Seen} & \textbf{Unseen} \\
  \midrule
  \#Scenes            & 51              & 36               & 2              & 44            & 4              \\
  \#Demonstrations    & 1048  & 91               & 63           & 182             & 101         \\
  \#Annotations       & 4691  & 412                & 302           & 867            & 493              \\
  \bottomrule
\end{tabular}
\captionof{table}{\textbf{Dynamic Object Affordance Task Split} 
Tasks where object usability may change during execution. These represent realistic scenarios requiring agents to infer preconditions and adapt. }
\label{tab:dynamic_tasks_distribution}
\end{minipage}

\begin{figure*}[t]
\centering
\begin{minipage}{0.8\textwidth}
\begin{lstlisting}[label={lst:pddl-domain}, caption={\textbf{Dynamic affordance scenario for tableware in a Stack and Place task.} This PDDL goal specification represents a scenario where dynamic object affordances occur on tableware items. }, captionpos=b]
(:goal
    (and
        (exists (?mo # object)
            (and
                (objectType ?mo {mrecep}Type)
                (isReceptacleObject ?mo)
                (cleanable ?mo)
                (isClean ?mo) 
            )
        )
        (exists (?mo # object)
            (and
                (objectType ?mo {mrecep}Type)
                (exists (?r # receptacle)
                    (and
                        (receptacleType ?r {recep}Type)
                        (exists (?o # object)
                            (and 
                                (objectType ?o {obj}Type)
                                (inReceptacleObject ?o ?mo)  
                                (inReceptacle ?mo ?r)      
                            )
                        )
                    )
                )
            )
        )
        (forall (?re # receptacle)
            (not (opened ?re))
        )
    )
)
\end{lstlisting}
\end{minipage}
\end{figure*}

\begin{table*}[t]
  \centering
  \small 
  \setlength{\tabcolsep}{4pt}  
  \resizebox{1.0\textwidth}{!}{
  \begin{tabular}{l
                  S[table-format=2.2] S[table-format=2.2]
                  S[table-format=2.2] S[table-format=2.2]
                  S[table-format=2.2] S[table-format=2.2]
                  S[table-format=2.2] S[table-format=2.2]}
    \toprule
    \multirow{2}{*}{\textbf{Method}} &
    \multicolumn{4}{c}{\textbf{Test Seen}} &
    \multicolumn{4}{c}{\textbf{Test Unseen}} \\
    \cmidrule(lr){2-5} \cmidrule(lr){6-9}
    & {\textbf{GC~$\uparrow$}} & {\textbf{PLW GC~$\uparrow$}} &
      {\textbf{SR~$\uparrow$}} & {\textbf{PLW SR~$\uparrow$}} &
      {\textbf{GC~$\uparrow$}} & {\textbf{PLW GC~$\uparrow$}} &
      {\textbf{SR~$\uparrow$}} & {\textbf{PLW SR~$\uparrow$}} \\
    \midrule
    \textbf{Few-Shot Methods} & & & & & & & & \\
    \midrule
    SayCan  & 21.15 & 15.00 & 2.34 & 0.58 & 21.90 & 12.73 & 0.00 & 0.00 \\
    LLM-Planner   & 26.10 & 7.49 & 4.21 & 1.89 & 28.33 & 8.47 & 4.77 & 3.08 \\
    \midrule
    \textbf{Supervised Methods} & & & & & & & & \\
    \midrule
    MOCA   & 19.50 & 17.17 & 2.33 & 0.59 & 21.47 & 17.03 & 0.00 & 0.00 \\
    FILM & 34.34 & 33.67 & 13.55 & 7.39 & 42.02 & 37.11 & 16.73 & 5.44 \\
    CAPEAM    & \textbf{55.08} & \textbf{43.82} & 37.85 & 18.65 & 58.84 & 43.16 & 37.05 & 14.18 \\
    \midrule
    \textbf{FILM + ADAPT (finetuned LLaVA)} & 31.86 & 33.52 & 10.74 & 5.58 & 46.21 & 44.99 & 21.51 & 7.87 \\
    \textbf{CAPEAM + ADAPT (finetuned LLaVA)} & 54.67 & 43.50 & \textbf{38.31} & \textbf{21.48} & 59.10 & \textbf{53.75} & \textbf{38.24} & \textbf{19.89} \\
    \textbf{CAPEAM + ADAPT (GPT-4o)} & 54.08 & 43.32 & 35.04 & 19.77 & \textbf{59.64} & 46.29 & 37.25 & 18.25 \\
    \bottomrule
  \end{tabular}
  }
  \caption{Main results on the DynAfford benchmark. (Tasks with Static Object Affordance)}
  \label{tab:static_results}
\end{table*}

\begin{table*}[t]
  \centering
  \small 
  \setlength{\tabcolsep}{4pt}  
  \resizebox{1.0\textwidth}{!}{
  \begin{tabular}{l
                  S[table-format=2.2] S[table-format=2.2]
                  S[table-format=2.2] S[table-format=2.2]
                  S[table-format=2.2] S[table-format=2.2]
                  S[table-format=2.2] S[table-format=2.2]}
    \toprule
    \multirow{2}{*}{\textbf{Method}} &
    \multicolumn{4}{c}{\textbf{Test Seen}} & 
    \multicolumn{4}{c}{\textbf{Test Unseen}} \\
    \cmidrule(lr){2-5} \cmidrule(lr){6-9}
    & {\textbf{GC~$\uparrow$}} & {\textbf{PLW GC~$\uparrow$}} &
      {\textbf{SR~$\uparrow$}} & {\textbf{PLW SR~$\uparrow$}} &
      {\textbf{GC~$\uparrow$}} & {\textbf{PLW GC~$\uparrow$}} &
      {\textbf{SR~$\uparrow$}} & {\textbf{PLW SR~$\uparrow$}} \\
    \midrule
    \textbf{Few-Shot Methods} & & & & & & & & \\
    \midrule
    SayCan  & 1.11 & 0.84 & 0.00 & 0.00 & 0.96 & 0.67 & 0.00 & 0.00 \\
    LLM-Planner   & 2.90 & 0.73 & 0.35 & 0.27 & 4.47 & 1.21 & 0.00 & 0.00 \\
    \midrule
    \textbf{Supervised Methods} & & & & & & & & \\
    \midrule
    MOCA   & 0.64 & 0.35 & 0.00 & 0.00 & 0.31 & 0.50 & 0.00 & 0.00 \\
    FILM   & 6.20 & 6.09 & 0.11 & 0.06 & 12.35 & 10.97 & 1.82 & 1.11 \\
    CAPEAM   & 12.24 & 8.06 & 2.19 & 1.50 & 18.28 & 19.20 & 1.41 & 1.35 \\
    \midrule
    \textbf{FILM + ADAPT (finetuned LLaVA)} & 12.65 & 9.77 & 3.11 & 0.98 & \textbf{24.97} & 18.37 & \textbf{10.75} & \textbf{3.47} \\
    \textbf{CAPEAM + ADAPT (finetuned LLaVA)} & \textbf{15.82} &\textbf{13.84} & \textbf{4.03} & \textbf{4.10} & 20.07 & \textbf{21.06} & 3.65 & 1.94 \\
    \textbf{CAPEAM + ADAPT (GPT-4o)} & 6.01 & 5.40 & 2.42 & 1.90 & 10.86 & 12.27 & 1.41 & 1.46 \\
    \bottomrule
  \end{tabular}
  }
  \caption{Main results on the DynAfford benchmark. (Tasks with Dynamic Object Affordance)}
  \label{tab:dynamic_results}
\end{table*}

\end{document}